# Computational Solar Energy - Ensemble Learning Methods for Prediction of Solar Power Generation based on Meteorological Parameters in Eastern India

DEBOJYOTI CHAKRABORTY & JAYEETA MONDAL, Department of Data Science and Engineering, BITS- Pilani, Rajasthan 333031, India

HRISHAV BAKUL BARUA, Robotics & Autonomous Systems, TCS Research, Kolkata 700156, India

ANKUR BHATTACHARJEE, Department of Electrical and Electronics Engineering, BITS-Pilani, Hyderabad Campus, Hyderabad, 500078, India

Corresponding Author:  Email id- a.bhattacharjee@hyderabad.bits-pilani.ac.in

**Abstract**

The challenges in applications of solar energy lies in its intermittency and dependency on meteorological parameters such as; solar radiation, ambient temperature, rainfall, wind-speed etc., and many other physical parameters like dust accumulation etc. Hence, it is important to estimate the amount of solar photovoltaic (PV) power generation for a specific geographical location. Machine learning (ML) models have gained importance and are widely used for prediction of solar power plant performance. In this paper, the impact of weather parameters on solar PV power generation is estimated by several Ensemble ML (EML) models like Bagging, Boosting, Stacking, and Voting for the first time. The performance of chosen ML algorithms is validated by field dataset of a 10kWp solar PV power plant in Eastern India region. Furthermore, a complete test-bed framework has been designed for data mining as well as to select appropriate learning models. It also supports feature selection and reduction for dataset to reduce space and time complexity of the learning models. The results demonstrate greater prediction accuracy of around 96% for Stacking and Voting EML models. The proposed work is a generalized one and can be very useful for predicting the performance of large-scale solar PV power plants also.

The official source code and implementation of this project can be found at our *project page in GitHub*

| Nomenclature | | |
|---|---|---|
| **Abbreviation** | **Full Form** | **SI Unit** |
| RH_Avg | Relative Humidity Average | RH |
| RA_mm | Rain Accumulation | millimeter |
| AT_Avg | Air Temperature Average | $°C$ |
| WS_Avg | Wind Speed Average | m/s |
| WD_Avg | Wind Direction Average | degree |
| DP_Avg | Dew Point Average | C |
| GR_Avg | Global Radiation Average | kWh/m2 |
| DiffR_Avg | Diffuse Radiation Average | kWh/m2 |
| AP_QNH_Avg | Atmospheric Pressure QNH Average | millibars |
| AP_QFE_Avg | Atmospheric Pressure QFE Average | millibars |
| DirR_Avg | Direct Radiation Average | kWh/m2 |
| HSR_Avg | Horizontal Solar Radiation Average | kWh/m2 |

Authors' addresses: Debojyoti Chakraborty & Jayeeta Mondal, Department of Data Science and Engineering, BITS-Pilani, Rajasthan 333031, India, debojyotichakrabarti@gmail.com,jayeetamondal5555@gmail.com; Hrishav Bakul Barua, Robotics & Autonomous Systems, TCS Research, Kolkata 700156, India, hbarua@acm.org; Ankur Bhattacharjee, Department of Electrical and Electronics Engineering, BITS-Pilani, Hyderabad Campus, Hyderabad, 500078, India, a.bhattacharjee@hyderabad.bits-pilani.ac.in.

| ISR_Avg | Inclined Solar Radiation Average | kWh/m2 |
|---|---|---|
| SA_Avg | Solar Azimuth Average | degrees |
| DTR_Avg | Direct Theoretical Radiation Average | kWh/m2 |
| GR_Avg | Global Energy Average | Joule |
| DiffE_Avg | Diffuse Energy Average | kWh/m2 |
| DirE_Avg | Direct Energy Average | kWh/m2 |
| ML | Machine Learning | - |
| EML | Ensemble Machine Learning | - |
| PV | Photovoltaic | - |
| GW | Giga watts | - |
| KW | Kilo watts | - |
| ARIMA | Auto-regressive integrated moving average | - |
| ANN | Artificial neural network | - |
| KNN | K-nearest neighbor | - |
| PCA | Principal component analysis | - |
| NMP | Numerical meteorological prediction | - |
| SVM | Support vector machine | - |
| GEFS | Global ensemble forecast system | - |
| SRRA | Solar radiation resource assessment site | - |
| DST | Department of Science and Technology | - |
| CEGESS | Centre of Excellence for Green Energy and Sensor Systems | - |
| KDE | Kernel density estimate | - |
| RMSE | Root mean squared error | - |
| GOSS | Gradient based one side sampling | - |
| EFB | Exclusive feature bundling | - |
| GBDT | Gradient boosting decision trees | - |
| GBM | Gradient boost machine | - |
| $R^2$ | R-Square score | - |
| ROI | Return on Investment | - |
| MNRE | Ministry of Renewable Energy | - |
| $R^d$ | Set of d dimensional real numbers | - |
| R | Set of real numbers | - |
| X | Set of input data points | - |
| Y | Set of ground truth data points for X input data points | - |
| E | Expected loss | - |
| n | Total number of input data points | - |
| D | Distribution over which an error is minimized | - |
| B | Parameter to update weight vector in Adaboost | - |
| $\psi$ | Expected loss over joint distribution in gradient boosting machine | - |
| L | Learning set for K regressors in | - |
| $\Omega$ | Penalty function for complexity of XGBoost model | - |

## 1 INTRODUCTION

To attenuate the impact of using fossil fuel-based sources of energy on the environment and to meet the ever-increasing power demands of the human society, industry, etc., renewable sources of energy are becoming the promising solutions. The annual share of electric power generation in the U.S. from renewable energy sources will rise from 20% in 2021 to 24% in 2023, as a result of continuing increases in solar and wind generating capacity [1]. India had 101.53 Gigawatts (GW) of renewable energy capacity in September 2021 which represented 38% of the overall installed power capacity. There is a current target of reaching about 450 GW of installed renewable energy capacity by 2030, where about 280 GW (over 60%) is expected to be from solar energy [2] itself. Considering these statistics and expected increase of solar power projects in the near future, it becomes essentially evident that accurate and efficient prediction of solar power generation [3] will be of high demand. In this paper, a practical case study, prediction and validation have been done for a kW scale solar PV power plant performance, installed in the Eastern region of India. This will have a great importance in solar PV Green-Field projects in terms of - 1) meteorological and environmental impact, 2) meeting solar generation expectations in a region, 3) suitable prediction algorithm, and 4) estimation of the return of investment (ROI)to solar investors. For smooth operation of power generation systems with considerably high solar power penetration,it is crucial to utilize a suitable solar power prediction scheme. In this paper, regional solar power prediction has been targeted with special attention to Eastern India data. Although there are a multitude of prediction methodologies, ensemble machine learning (EML) based prediction algorithms have been focused in this work, given their large-scale success in forming accurate systems. Since, in general meteorological data can be time series in nature learning based methods can be employed in some way or another [4].

There exists significant number of earlier works on deep learning based solar power prediction systems [5], [6], [7], [8],[9]. Popular data analysis tools like principal component analysis (PCA) to construct suitable dataset for multivariate analysis of solar irradiance prediction are adopted in some earlier works [10], [11], [12]. Some earlier works show using domain knowledge based deep learning method [13], short term predictive models [14], Pearson's correlation based prediction [15], prediction with adjusting the site-specific photovoltaic (PV) output forecast [16] to predict the solar radiation. Authors in [17], [18], [19], have used multiple predictive regression models and ensemble techniques to improve the accuracy of solar power prediction. Here a short but comprehensive survey has been put forward on the related works in this domain from the perspective of our paper's contributions. By conducting a thorough earlier work study, it has been concluded that data mining techniques play an important role in significantly improving the accuracy of any ML based prediction system. Voyant *et al.* [50] proposed a robust functional model based on prediction intervals approach with efficiency and interest better than classical prediction tools based on bootstrap utilization. But This kind of model is dedicated to periodic time series and is not restricted to stationary time series. Ayodele *et al.* [51] implemented a k-means hybridized with support vector machine (SVR) for global solar radiation prediction. The author used k-means to cluster seasonal data into clusters with centroids for wet and dry seasons. The proficiency of the model was verified using six years of meteorological data of Ibadan (2010–2015) to predict daily global solar radiation for the future years (2016–2017). The predicted values were also compared with the forecasted values obtained from three established models i.e. ANN, Angstrom–Prescott and ARMA models. Peder *et al.*and Coimbra *et al.* [20] conducted a study for hourly values of solar power for horizons of up to 36 hours where auto-regressive integrated moving average (ARIMA), artificial neural network (ANN) and k-nearest neighbor (KNN) models were bench-marked and found. Historical output levels ofthe panels were used, i.e., no exogenous numerical meteorological prediction (NWP) data was used as input. Davò *et al.* [21] used PCA technique as a feature selection method for dimensionality reduction and implemented ANN and analog ensemble (AnEn) to predict solar irradiance. Authors claimed that minor variations in daily meteorological depending on cloud coverage, humidity level, presence of fog, etc., can affect the solar power generation. Shi *et al.* [22] proposed an algorithm for meteorological classification and support vector machine (SVM) to forecast PV power outputon 15-minute intervals for the fore-coming day. The meteorological data is classified into four classes: clear sky, cloudyday, foggy day, and rainy day. In [23] the authors have prepared a study on NWP computed from GEFS, the global ensemble forecast system, which predicts meteorological variables for points in a grid. They have performed a study ofdifferent ML techniques in the context of solar energy forecasting using NWP models computed from the NOAA/ESRLGEFS for different nodes located in a grid. In our study, the variety of ML algorithms for training and testing on our meteorological data has been further increased to better understand the variation and prediction analysis results. A comparison between existing similar works and our paper is presented systematically in Table. 1.

Table 1. Comparison of existing works similar to ours (considering last 5 years).

| Works | Methods used | Predicted parameter | Ensemble Learning used | Field Data testing | End-to-End Data Analysis and ML Framework (Extensive meteorologicalfeature engineering and analysis done |
|---|---|---|---|---|---|
| Fan et al. (2018) [24] | SVM and XGBoost | Daily globalsolar radiation | ✗ | ✗ | ✗ |
| Srivastava et al. (2018) [25] | Feed-forward, back propagation, DL, and model averaging networks | 6 day-a-headforecasting global solarradiation | ✗ | ✓ | ✗ |
| Mpfumali et al. (2019) [26] | Stochastic gradient boosting and support vector regression | Day aheadhourly global horizontal irradiance | ✗ | ✓ | ✗ |
| Coban et al.(2019) [27] | Classical ML models, probabilistic naive models, and tree-based regression | Global horizontal irradiation | ✗ | ✗ | ✗ |
| Chen et al.(2020) [28] | Static and dynamic ensembles | Power generated | ✓ | ✓ | ✗ |
| Amarasinghe et al.(2020) [29] | Stacking and weighted averaging ensemble methods using Deep Belief Network (DBN), random forest (RF), and SVM | Power generated | ✓ | ✓ | ✗ |
| Mutavhatsindi et al. (2020) [30] | Static and dynamic ensembles | Power generated | ✓ | ✓ | ✗ |
| Guermoui et al. (2020) [31] | General ensemble learning approach (GELA), cluster-based ensemble learning approach (CELA), decomposition-based ensemble learning approach (DELA), decomposition clustering based learning approach (DCLE), evolutionary ensemble learning approach (EELA) | Global horizontal irradiation | ✓ | ✗ | ✗ |
| Zhou et al.(2021) [32] | Review of ANN, kernel, tree, ensemble, decomposition cluster-based ML algorithms | Global solarradiation | ✓ | ✗ | ✗ |
| Jebli et al.(2021) [33] | Recurrent neural network (RNN), Long Short Term Memory (LSTM), and Gated Recurrent Unit (GRU) | Solar energy | ✗ | ✓ | ✗ |
| **Proposed work** | **Classical ML algorithm and ensemble techniques (bagging, boosting, stacking, and voting)** | **SolarPower** | ✓ | ✓ | ✓ |

Based on the above discussion, the major contributions in this paper are represented in a threefold manner.

1. The collected field data from the Solar Radiation Resource Assessment (SRRA) center and the rooftop solarPV power plant has been pre-processed (data cleaning, dimensionality reduction and feature selection) andanalyzed.

2. The processed dataset has been fed to various ML models for prediction of solar PV power generation under the impact of meteorological parameters (direct, global, inclined and diffused radiation, ambient temperature, solar azimuth, wind speed, wind direction and relative humidity). At this stage it has been decided that the EML models are superior in terms of prediction accuracy compared to the other classical regression models. The EML models chosen in this work are Bagging, Boosting, Stacking and Voting.

3. The performance of the algorithms is validated by a $10kW_p$ rooftop solar power plant dataset. It is found from the prediction results that the Stacking and Voting EML algorithms demonstrates prediction accuracy of around 96%, which is better than the other EML algorithms considered in this work. Thus, the recommended EML algorithms can be very useful for predicting performance of even large-scale solar power plants subject to specified geographical location. Further, it can be inferred that the proposed test-bed can also be utilized in any green-field solar project for predicting its performance with proper identification of suitable training parameters.

*The official working code (along with the pre-trained learning models and the novel meteorological dataset) of the proposed test-bed framework have been released at our **project web page in GitHub**.*

The paper is organized as follows. Our proposed methodology of study and analysis has been described in Section 2. The dataset collection process and setup, pre-processing, analysis and the impact of several meteorological parameters on solar power generation is studied is elaborated in Section 3. The various learning techniques used in the papers are explored in Section 4.1, thereafter experimentation has been discussed in Section 4.2. Analysis of the result is shown in Section 4.3 followed by conclusion of our analysis, studies, and experimentation in Section 5.

## 2 PROPOSED FRAMEWORK OF OUR STUDY AND ANALYSIS: OVERVIEW AND DESIGN

In this section, the proposed test-bed framework for our study, analysis, and experimentation has been summarized. The flow begins by collection and curation of data from the source at SRRA site. Since in this paper, the SRRA meteorological data has been used that was collected at the regional power generation station, it suffers from issues like missing value and outliers due to sensor system shutdowns or sensor faults etc. Hence, the pipeline has to be carefully designed so that model predictions are less affected due to the mentioned quality issues in the collected data. The block diagram of the framework (with the data/information flow arrows) has been shown in Figure 1.

(1) ***Data Acquisition***: Section 3.1 discusses the collection of field dataset used in this work. Use of field data helps in validating claims, which can be reproduced when Greenfield solar projects are implemented.

(2) ***Data Analysis and Prepossessing***: Once data is collected, it requires cleaning to remove errors in dataset which has been discussed in Section 3.2.1. Exploratory data analysis bright insights about the impact of features of the meteorological data towards predicting the target variable. The various steps taken for analysis and processing are shown in Section 3.2.2.

(3) ***Ensemble Learning Models Exploration, Experimentation and Result Analysis***: The section has been divided as-

    I. In Section 4.1 the various ensemble learning models has been explored theoretically.

    II. Followed by Section 4.2 the experimental setup and process for training and testing the EML algorithms has been shown.

    III. Finally, in Section 4.3 compare the various algorithms tested on SRRA dataset based on evaluation metric and performance. Algorithms best suited for eastern region meteorological data are proposed.

## 3 PRACTICAL DATA ACQUISITION, PRE-PROCESSING, AND DATA ANALYSIS

This section of the paper discusses the meteorological dataset collected in the Eastern Region of India and the various methodologies adopted for data analysis and pre-processing.

### 3.1 Data Acquisition

The solar power generation dataset used in this work consists of consequent meteorological parameters and the corresponding power generated for each set of observations. Data has been collected from Solar Radiation Resource

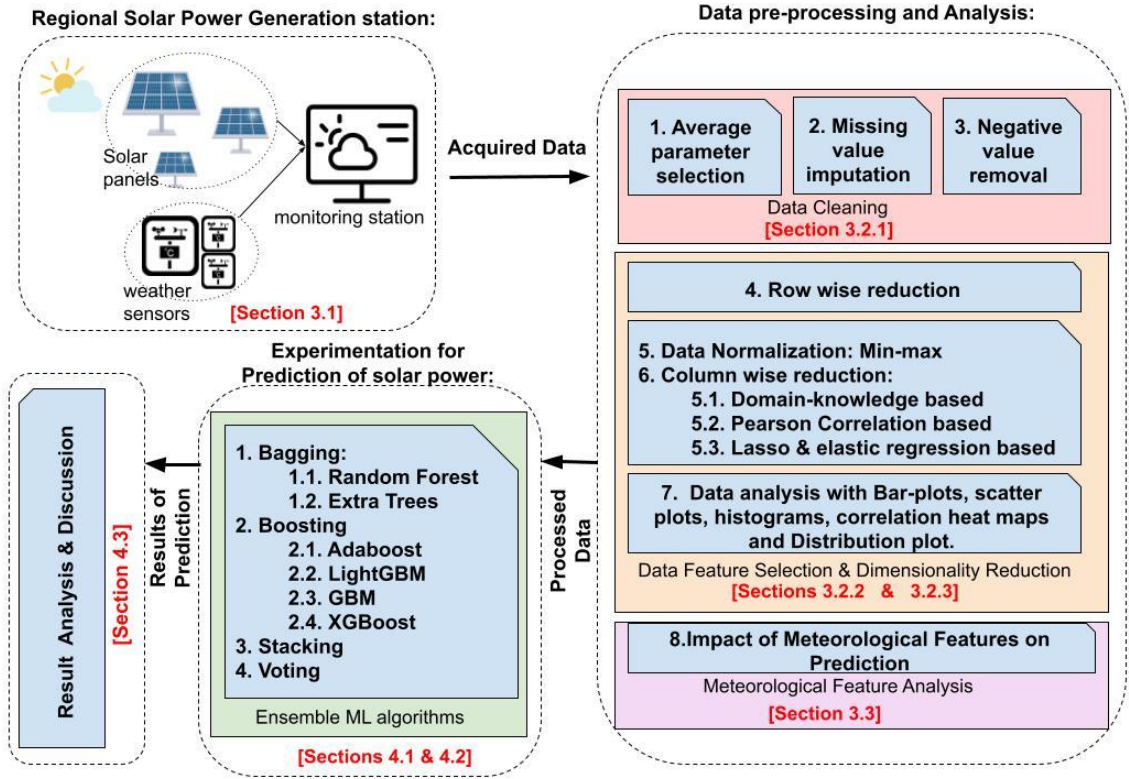

Fig. 1. The complete bird's eye view of the proposed Test bed Framework depicting the workflow for Data Collection/Curation, Data Mining/Analytics and Supervised Learning Methods study, analysis, and experimentation on SRRA regional meteorological data(here data from East India) for Solar power forecasting and prediction.

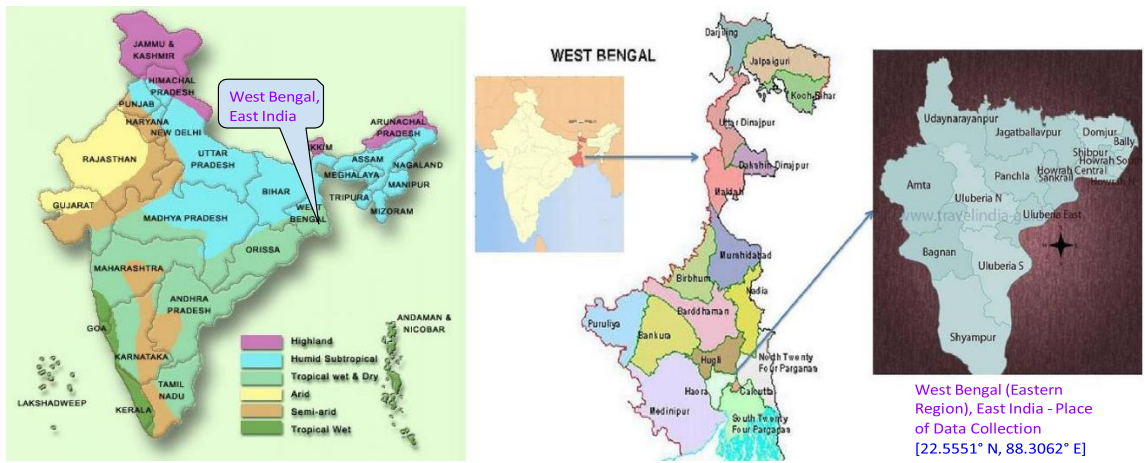

Fig. 2. The place (in West Bengal (WB), East India) for Data collection (depicted in map: 22.5551° N, 88.3062° E) with general meteorological conditions (Pictures collected from Google Images).

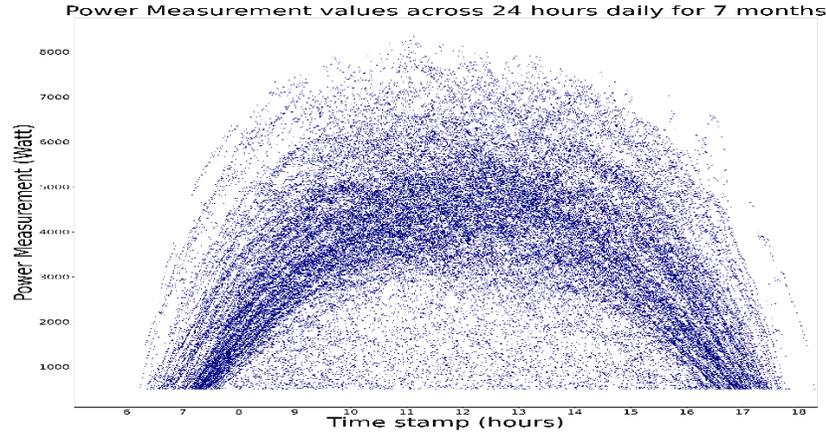

Fig. 3. Visualization of solar power generation measurement in 24 hours across 7 months duration.

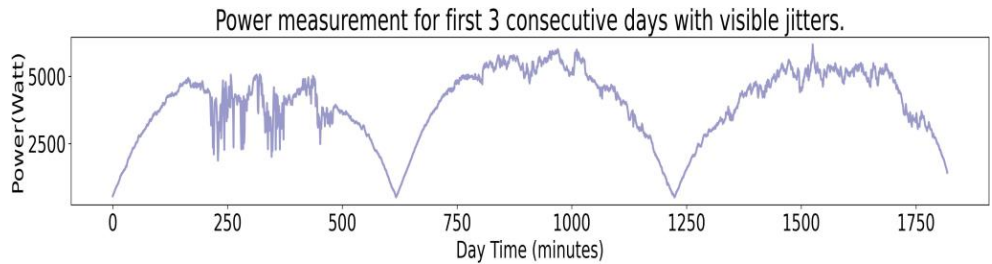

Fig. 4. Day-time power measurement of first 3 consecutive days to show visible jitters in solar power.

Assessment site (SRRA)[1] established (in 2014) and solar PV power plant in the roof top of Indian Institute of Engineering Science and Technology (IIEST), Shibpur, West Bengal (WB), East India shown in Figure 2. It is a part of DST-IIEST Solar Photovoltaic (PV) Hub[2] which happens to be a joint initiative by Department of Science and Technology (DST)[3], Government of India and Centre of Excellence for Green Energy and Sensor Systems (CEGESS)[4], Indian Institute of Engineering Science and Technology (IIEST), Shibpur Figure 5 a. The rooftop 10 kW$_p$ solar PV power plant measures the power output and the SRRA center collects meteorological parameters in real-time. The data collected was stored in batches at a centralized monitoring station over a duration of 7 months (August 2019 - February 2020). Figure 3 shows sample plots of power generation (in Watt) registered at 1 minute interval on four distinctive days carefully selected to visualize typical the variations in the curves. From the figure it is observed that the curve is shaped like a table-top between 6:00 hoursand 18:00 hours of day-time while the top of the curve representing the maximum power values at around mid-day,that is, approximately around 11:00 to 13:00 hours. Furthermore, in Figure 4 some random jitters can be observed in the curves, which can be a probable result of cloud formation and hindrance of solar power generation. The original dataset consists of 286187 number of data-points and 40 attributes including target attribute power. Table 2 describes the various attributes in the target dataset.

---

[1]http://dst-iiestsolarhub.org.in/about_SRRA.php
[2]http://dst-iiestsolarhub.org.in/about_DST_IIEST_solar_hub.php
[3]https://dst.gov.in/
[4]https://oldww.iiests.ac.in/index.php/home-g

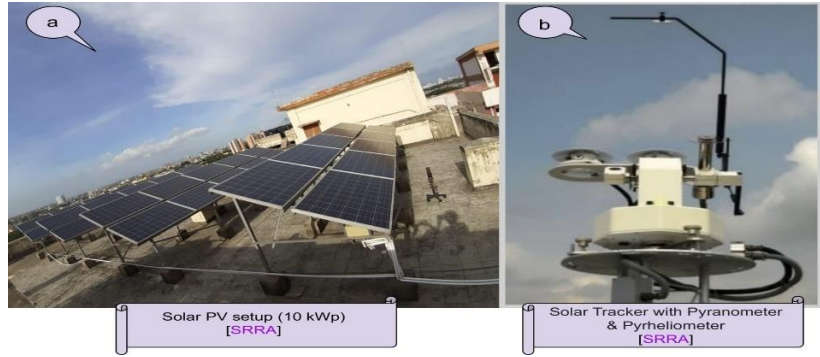

Fig. 5. Images from the Standard measurement system and actual experimental setup for our data collection/curation in SRRA, WestBengal, East India.

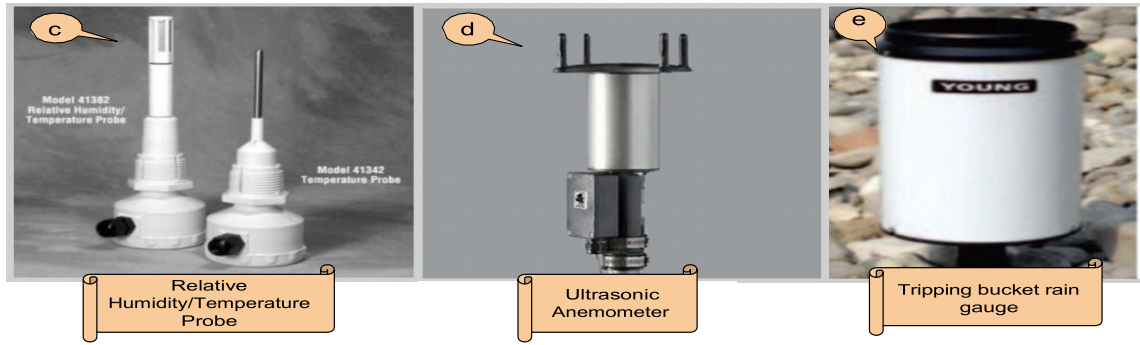

Fig. 6. Images of the Meteorological measurement instruments in SRRA which are used in our data collection/curation and pre- processing stage.

Table 2. Description Of Meteorological Features used.

| Parameter | Instruments used | Specifications | Manufacturer and Model |
|---|---|---|---|
| RH_Avg | Relative Humidity probe [Figure 6 c] | RELATIVE HUMIDITY: (41382)<br>Measuring Range: 0-100 %RH<br>Accuracy at 20 °C: ±2 %RH.<br>Stability: Better than ±1 %RH per year<br>Response Time: 10 seconds (without filter)<br>Sensor Type: Rotronic Hygromer TM<br>Output Signal: V option: 0-1 VDC,<br>L option: 4-20 mA | R.M. YOUNG COMPANY<br>Traverse City, Michigan 49686 USA |
| RA_mm | Tripping Bucket Rain Gauge [Figure 6 e] | Resolution: 0.1 mm per tip<br>Accuracy: 2% up to 25 mm/hr<br>3% up to 50 mm/hr<br>Output: Magnetic reed switch (normally open)<br>Contact rating: 24VAC/DC 500mA<br>Operating Temperature: -20 to +50℃ (heated)<br>Heater Thermostat Setpoint: 10℃ ±3℃ | R.M. YOUNG COMPANY<br>Traverse City, Michigan 49686 USA |
| AT_Avg | Temperature Probe [Figure 6 c] | TEMPERATURE: (41382, 41342)<br>Calibrated Measuring Range:<br>-50 to 50 °C (suffix C)<br>-50 to 150 °F (suffix F)<br>Response Time: 10 seconds (without filter)<br>Accuracy at 0 °C: ±0.3 °C**<br>±0.1 °C (optional) with NIST traceable calibration<br>Sensor Type: Platinum RTD<br>Output Signal: V Option: 0-1 VDC, L Option: 4-20 mA, 4 wire RTD (41342 only) | R.M. YOUNG COMPANY<br>Traverse City, Michigan 49686 USA |
| WS_Avg | Ultrasonic Anemometer [Figure 6 d] | Measuring range :0.01 – 75 m/s<br>Accuracy: ±2% of measured value for >5ms<br>Resolution: 0.01m/s | Biral, Bristol Industrial & Research Associates Ltd |
| WD_Avg | Ultrasonic Anemometer [Figure 6 d] | Measuring range: 0-360°<br>Resolution: 1°<br>Accuracy: ±1° | Biral, Bristol Industrial & Research Associates Ltd |
| DP_Avg | Hygrometer | Measurement ranges: 0 -100 %     RH- 30 -100 °C<br>Resolution: 0.01% RH 0.01 °C<br>Accuracy: ±2.0 % RH at 25 °C / 20 ... 80 % RH<br>other measurement areas: ±2,5 % RH<br>Measurement rate: 2,5 / second<br>Display: dual LCD with 4.5 positions (backlit)<br>Response time: RH<br>about 10 seconds | PCE Instruments UK Limited |
| GR_Avg | Pyranometer [Figure 5 b] | Sensitivity: approximately 9 $\mu$V/Wm$^{-2}$.<br>Impedance: approximately 650 Ohms.<br>Temperature Dependence: ±1% over ambient temperature range -20 to +40°C. Linearity: ±0.5% from 0 to 2800 Wm$^{-2}$.<br>Response time: 1 second (1/e signal).<br>Cosine Response: ±1% from normalization 0-70° zenith angle; ±3% 70-80° zenith angle. | Eppley lab, Rhode island, USA. PSP |
| DiffR_Avg | Sky Scanning Radiometer [Figure 7 g] | Measurement principle: Multi-band filter radiometer UV, Visible and NIR radiometer<br>Detector: UV-enhanced silicon photo-diode<br>Wavelengths: 315, 400, 500, 675, 870, 940 and 1020 nm<br>Wavelength accuracy: 2 nm | Kipp & Zonen B.V Delftechpark 36, 2628 XH Delft P.O. Box 507, 2600 AM Delft The Netherlands |
| AP_QNH_Avg | Digital Barometer | Pressure Range:500 to 1100 hPa<br>Operating Temperature:-40 to +60°C<br>Digital Accuracy*0.2 hPa (25°C)0.3 hPa (-40 to +60°C)<br>Analog Accuracy** 0.05% of analog pressure range<br>Analog Temperature Dependence<br>0.0017% of analog pressure range/ °C (25°C reference)<br>Update Rate:1.8 Hz max<br>Serial Output:Full duplex RS-232 9600 baud<br>Polled or continuous ASCII text, NMEA<br>Half duplex RS-485 (61302L only)<br>Analog Output: 0 to 5000 mV, 0 to 2500 mV (61302V)<br>4 to 20 mA (61302L)<br>Resolution: Serial 0.01 hPa Analog 0.025% of analog scan | R.M. YOUNG COMPANY Traverse City, Michigan 49686 USA |
| AP_QFE_Avg | Digital Barometer | Pressure Range:500 to 1100 hPa<br>Operating Temperature:-40 to +60°C<br>Digital Accuracy*0.2 hPa (25°C)0.3 hPa (-40 to +60°C)<br>Analog Accuracy** 0.05% of analog pressure range<br>Analog Temperature Dependence<br>0.0017% of analog pressure range/ °C (25°C reference)<br>Update Rate:1.8 Hz max<br>Serial Output:Full duplex RS-232 9600 baud<br>Polled or continuous ASCII text, NMEA<br>Half duplex RS-485 (61302L only)<br>Analog Output: 0 to 5000 mV, 0 to 2500 mV (61302V)<br>4 to 20 mA (61302L)<br>Resolution: Serial 0.01 hPa Analog 0.025% of analog scan | .M. YOUNG COMPANY Traverse City, Michigan 49686 USA |
| DirR_Avg | Multi Spectral Photometer [Figure 7 f] | Resolution: 0.05 degree<br>Maximum Speed: 10 degree/sec<br>Traverse: 280º<br>Actuator: Stepper Motor ( DPM60SH86 - 2008AF ) | Holmarc Opto-Mechatronics LTD, Kerala, India |
| HSR_Avg | Pyrgeometers [Figures 7 i & j] | Spectral range (overall)4.4 to 50 μm = 4400 to 50000 nm<br>Response time (95%)< 18 s<br>Window heating offset< 15 W/m²<br>Zero offset B< 5 W/m²<br>Temperature dependence of sensitivity (-10 ℃ to +40 ℃)< 3 %<br>Operational temperature range-40 to +80 °C<br>Field of view150 °<br>Non-linearity< 1 %<br>Analogue output (-A / -V version)4 to 20 mA / 0 to 1 V<br>Digital output2-wire RS-485 | OTT HydroMet B.V. Delft - The Netherlands |
| ISR_Avg | Pyrgeometers [Figures 7 i & j] | Spectral range (overall)4.4 to 50 μm = 4400 to 50000 nm<br>Response time (95%)< 18 s<br>Window heating offset< 15 W/m²<br>Zero offset B< 5 W/m² | OTT HydroMet B.V. Delft - The Netherlands |

| | | Temperature dependence of sensitivity (-10 ℃ to +40 ℃)< 3 %<br>Operational temperature range-40 to +80 °C<br>Field of view150 °<br>Non-linearity< 1 %<br>Analogue output (-A / -V version)4 to 20 mA / 0 to 1 V<br>Digital output2-wire RS-485 | |
|---|---|---|---|
| SA_Avg | Azimuth Compass | Resolution: 1Â°<br>Accuracy: 0.5Â° | Brunton International LLC, Riverton, Wyoming, US. |
| DTR_Avg | UV Radiometer [Figure 7 g] | Spectral range (overall)280 to 400 nm<br>Sensitivity300 - 500 µV/W/m²<br>Temperature response< -0.3% per °C<br>Response time (95%)< 1 s<br>Non-linearity (0 to 100 W/m²)< 1 %<br>Maximum UVA/UVB irradiance400 W/m²<br>Operational temperature range-40 to +80 °C<br>Directional response (up to 70° with 100 W/m² beam)< 5 %<br>Field of view180 ° | OTT HydroMet Corp. USA Sterling - USA |
| GR_Avg | Pyranometer [Figure 5 b] | Sensitivity: approximately 9 µV/Wm$^{-2}$.<br>Impedance: approximately 650 Ohms.<br>Temperature Dependence: ±1% over ambient temperature range -20 to +40°C. Linearity: ±0.5% from 0 to 2800 Wm$^{-2}$.<br>Response time: 1 second (1/e signal).<br>Cosine Response: ±1% from normalization 0-70° zenith angle; ±3% 70-80° zenith angle. | Eppley lab, Rhode island, USA. PSP |
| DiffE_Avg | Sky Scanning Radiometer [Figure 7 g] | Measurement principle: Multi-band filter radiometer UV, Visible and NIR radiometer<br>Detector: UV-enhanced silicon photo-diode<br>Wavelengths: 315, 400, 500, 675, 870, 940 and 1020 nm<br>Wavelength accuracy: 2 nm | Kipp & Zonen B.V Delftechpark 36, 2628 XH Delft P.O. Box 507, 2600 AM Delft The Netherlands |
| DirE_Avg | Multi Spectral Photometer [Figure 7 f] | Measurement principle: Multi-band filter radiometer UV, Visible and NIR radiometer<br>Detector: UV-enhanced silicon photo-diode<br>Wavelengths: 315, 400, 500, 675, 870, 940 and 1020 nm<br>Wavelength accuracy: 2 nm | Kipp & Zonen B.V Delftechpark 36, 2628 XH Delft P.O. Box 507, 2600 AM Delft The Netherlands |

### 3.2   Data Analysis and Pre-processing

The unprocessed sensor field dataset used in this work has been collected directly from source, which suffers from the issues related to sensor logging errors. The uncertainties in data recording may arise from instrument selection, instrument condition, instrument calibration, environment, observation and reading and test planning. Therefore, data cleaning and preprocessing is necessary before feeding them to the ML models for training. In this section of the paper, the details of various data mining techniques implemented, broadly classified under data cleaning and also carry out data distribution analysis, has been provided in order to retrieve meaning-full patterns from the dataset.

*3.2.1   Data Cleaning by Missing Value Imputation:* In cases of logs containing missing data (appears when there is a system glitch in the collection phases due to various issues like system's downtime), an easy solution of discarding them can cause data volume reduction and removal of significant data-points. Hence, a technique called missing value imputation [34] has been implemented to replace the missing values with mean imputation, regression imputation, hot deck imputation, cold deck imputation or median imputations. Based on the nature of our data mean value imputation has been implemented.

*3.2.2   Feature Selection and Dimensionality Reduction:* Increase in number of training features makes a predictive modelling task more challenging referred to as the curse of dimensionality. Dimensionality reduction refers to techniques that reduces the number of features in a dataset. Data reduction/dimensionality reduction is typically conducted in two directions, i.e., row-wise for data sample reduction and column-wise for data variable reduction. Since the active sun time for India is around 8 to 10 hours and a significant amount of target data is zero which creates a bias. This bias can be eliminated by row-wise data sampling, which has been performed by eliminating rows having target variable (power generated) less than a threshold value (500 Watt). In this manner the number of data-points have been reduced from 286187 to 92684 with 42 meteorological features. The three main methods of column-wise data variable reduction i) use domain knowledge to directly select variables of interest, ii) use statistical feature selection methods to select important variables for further analysis, iii) use Pearson's correlation to find highly correlated features and drop one among the two. For column-wise data reduction only average value readings of meteorological features were selected and instantaneous, max and min data readings of meteorological features were discarded. Certain sensor data like "Pyrheliometer error" and "Geotrac3K Status" were also discarded since they do not affect the use case. Hence the column dimension has been reduced from 40 to 18 features. Thereafter an L1 regularization method called lasso regression was used to find features of highest importance. The lasso method regularizes model parameters by shrinking the regression coefficients, reducing some of them to zero. The feature selection phase occurs after the shrinkage, where every non-zero value is selected to be used in the model. This method is significant in the minimization of prediction errors that are common in statistical models. To give confidence to feature selection elastic net regression was used with L1 and L2 penalties during training, resulting in better performance than a model with either one or the other penalty on some problems. Figure 8 shows the importance plot using both the methods discussed above. In Table 3 coefficients of the two regularization methods have been listed.

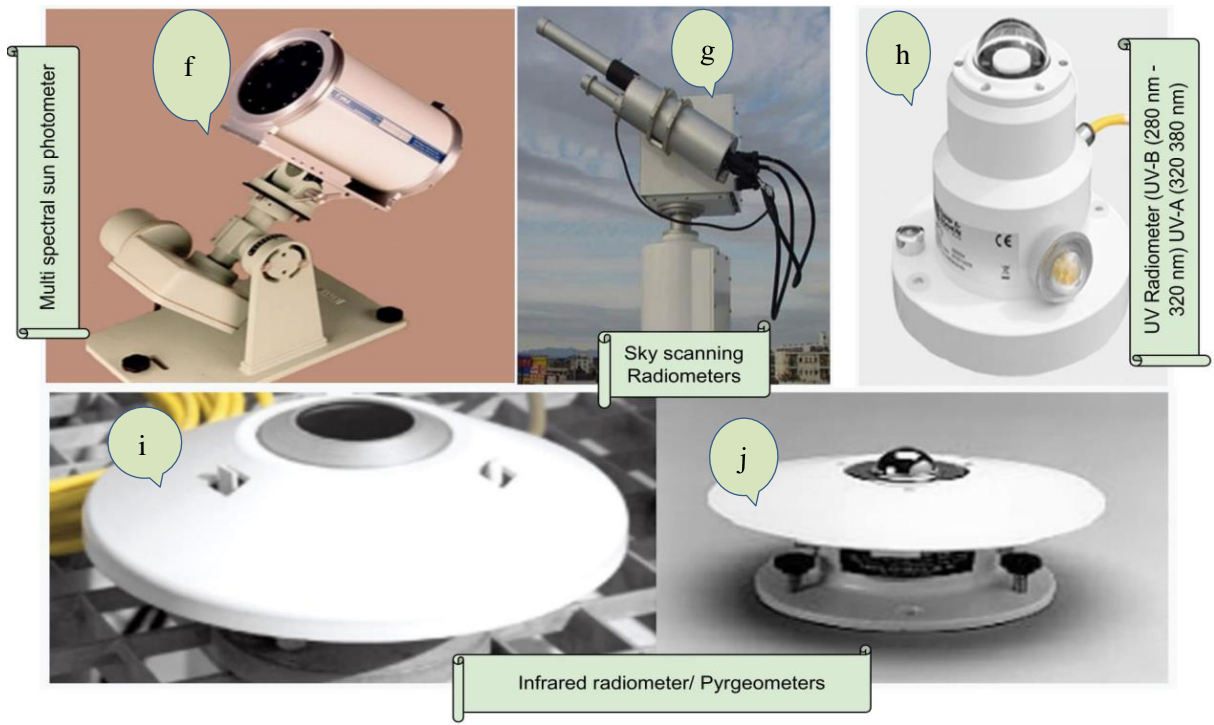

Fig. 7. Images of the other Advanced measuring equipments from SRRA which are used in our data collection/curation process.

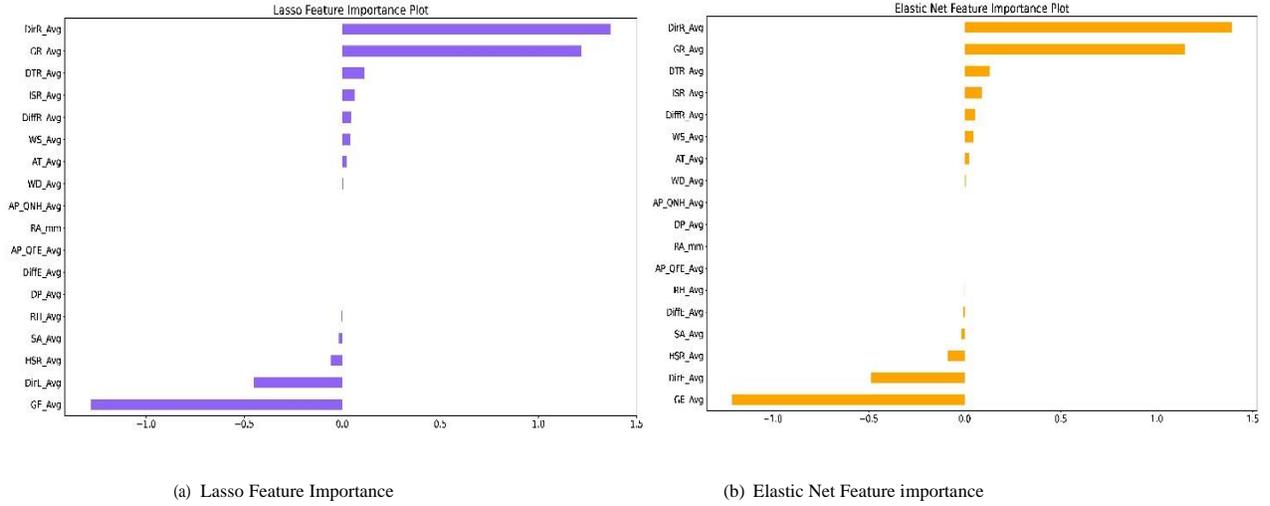

(a) Lasso Feature Importance

(b) Elastic Net Feature importance

Fig. 8. Feature importance Plot for Feature Selection.

Hence the common least important features 'AP_QFE_Avg', 'RA_mm', 'DP_Avg', and 'AP_QNH_Avg' were eliminated. Although theoretically rainfall is an indirect measure of cloud cover inturn affecting power generation, in our case it was found that the rainfall accumulation has a low feature importance. Tocross verify the logic variance of RA_mm has been checked to be very low as 3.36 x 10^-5. So, there could be a possibility of incorrect data collection by rainfall accumulation sensor. Hence it was discarded as it is going to affect the accuracy of our prediction. Further analysis from finding Pearson's correlations [35] between the various attributes by visualizinga heat-map of Pearson's coefficient, as shown in Figure 9(a). From the given plot the highly correlated attributes have been segregated with a coefficient threshold of 0.95. Some of the highly correlated pairs include: HSR_Avg and ISR_Avg, GR_Avg and GE_Avg, DiffR_Avg and DiffE_Avg, DirR_Avg and DirE_Avg, DTR_Avg and DirR_Avg, etc., which have been removed from the dataset in order to prepare a better input feature space for machine learning algorithms as per observations from Figure 9(a). The final set of features include 'RH_Avg', 'AT_Avg', 'WS_Avg', 'WD_Avg', 'GR_Avg', 'DiffR_Avg', 'DirR_Avg', 'ISR_Avg', and 'SA_Avg'. After dimensionality reduction the correlation coefficients of these input features are relatively low as shown in Figure 9(b). Hence a reduction of dimension has been observed from 18 trainable parameters to 9.

Table 3. Feature Importance table for Feature Selection.

| Sr. No. | Feature | Lasso Importance | E-Net Importance |
|---|---|---|---|
| 1 | GE_Avg | -1.281 | -1.212 |
| 2 | DirE_Avg | -0.449 | -0.491 |
| 3 | RH_Avg | -0.002 | -0.002 |
| 4 | SA_Avg | -0.017 | -0.018 |
| 5 | WD_Avg | 0.005 | 0.005 |
| 6 | DP_Avg | 0.000 | 0.000 |
| 7 | AP_QFE_Avg | 0.000 | 0.000 |
| 8 | HSR_Avg | -0.060 | -0.0885 |
| 9 | DTR_Avg | 0.111 | 0.130 |
| 10 | RA_mm | 0.000 | 0.000 |
| 11 | DiffE_Avg | 0.000 | -0.007 |
| 12 | AT_Avg | 0.021 | 0.023 |
| 13 | AP_QNH_Avg | 0.000 | 0.000 |
| 14 | WS_Avg | 0.043 | 0.043 |
| 15 | ISR_Avg | 0.064 | 0.089 |
| 16 | DiffR_Avg | 0.046 | 0.054 |
| 17 | DirR_Avg | 1.369 | 1.391 |
| 18 | GR_Avg | 1.219 | 1.147 |

Some of the highly correlated pairs include: HSR_Avg and ISR_Avg, GR_Avg and GE_Avg, DiffR_Avg and DiffE_Avg, DirR_Avg and DirE_Avg, DTR_Avg and DirR_Avg, etc., which have been removed from the dataset in order to prepare a better input feature space for machine learning algorithms as per observations from Figure 9(a). The final set of features include 'RH_Avg', 'AT_Avg', 'WS_Avg', 'WD_Avg', 'GR_Avg', 'DiffR_Avg', 'DirR_Avg', 'ISR_Avg', and 'SA_Avg'. After dimensionality reduction the correlation coefficients of these input features are relatively low as shown in Figure 9(b). Hence a reduction of dimension has been observed from 18 trainable parameters to 9.

*3.2.3  Kernel Density Estimate (KDE) analysis :* In this subsection Kernel density estimation is the process of estimating an unknown probability density function using a kernel function. Kernel density estimation is a non-parametric method of estimating the probability density function (PDF) of a continuous random variable. With reference to Figure 10 the following are observed-

i) Unimodal distribution is followed by 'Power', 'RH_Avg', 'WS_Avg', 'WD_Avg', 'DiffR_Avg', 'DirR_Avg', and 'GR_Avg'.

ii) Bimodal distribution is followed by ISR_Avg, SA_Avg. This indicates that their can be two different kind of groups in bimodal distributions will need the regression model must be able to take care of these distributions separately.

### 3.3  Impact of meteorological parameters on Solar PV power generation

Solar power plant's performance depends on meteorological parameters and hence the power generation is intermittent in nature. In this work, major environmental parameters which directly influence PV power generation have been considered. To validate our analysis of features and their importance in predictions, it has been shown that the power generation calculated using certain environmental parameters by [36] is close to actual power generated. The various

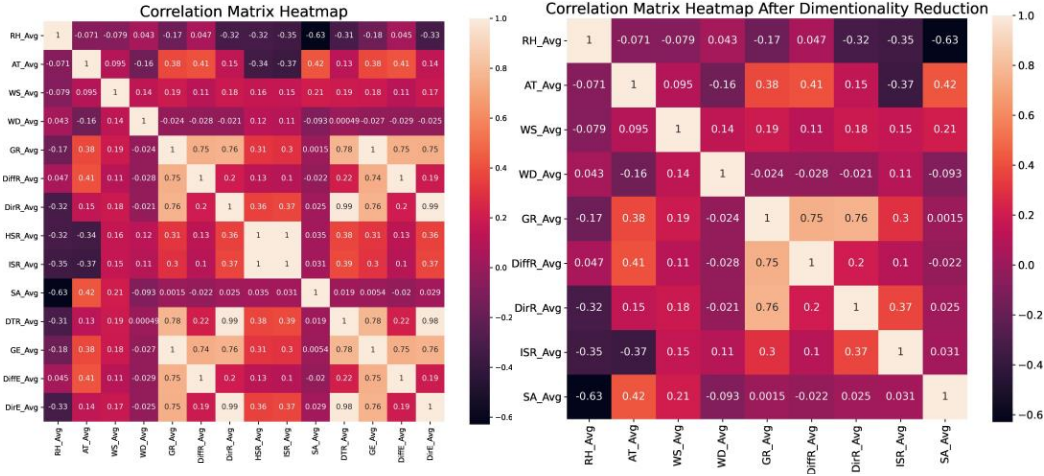

(a)  Correlation Plot before feature reduction          (b)  Correlation Plot after dimensionality reduction

Fig. 9.  Pearson's Correlation Coefficient Plot for dimensionality reduction.

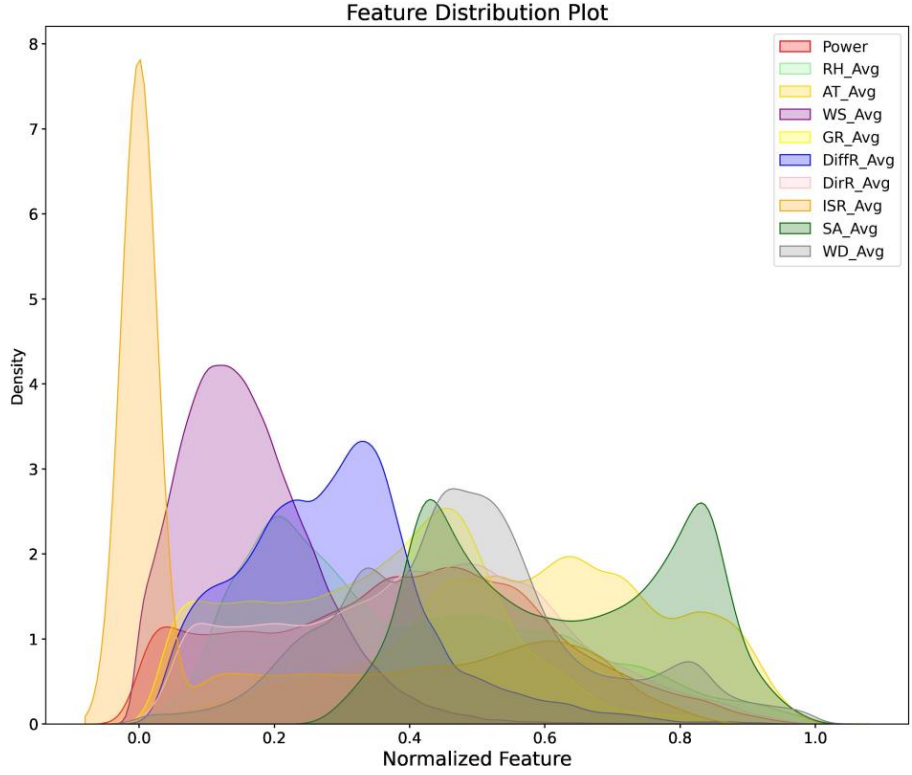

Fig. 10. Data Distribution Plot.

equations used to calculate the power are Equations 1- 4. The definition and values of the constants used are given in Table 4 and Table 5. In Table 6, a random sample range of 20 observations of power generation calculated using DR_Avg and WS_Avg and actual power has been tabulated. Using this method [36] a root mean squared error (RMSE) of 799.25 is observed against the actual power generated. Hence choice of environmental parameters heavily affects the results. But it has been shown that DR_Avg and WS_Avg are not the only features that are responsible but inclusion of fewmore parameters improves our predictions.

$$S_h = S_l \sin \propto \tag{1}$$

$$S_m = S_l \sin(\propto + \beta) \tag{2}$$

$$T_i = S_m/(26.9 + (6.2 * wi) \tag{3}$$

$$P_c = P_r S_m \big(1 + (k1 \log(S_m)) + k2 \log(S_m{}^2) + k3 T_i\big) + k4 T_i \log(S_m) + k5 T_i \log(S_m{}^2) + k6 T_i{}^2 \tag{4}$$

From surveying several literature papers on solar radiation and solar power generation prediction analysis, it has been observed that solar azimuth and ambient temperature parameters play a direct role in solar radiation measurementand eventually the amount of solar power generated. Figure 11 shows the 3-Dimensional plot representing how power generation varies with average normalized values of ambient temperature and solar azimuth measure. It is a scatter plot on 3-D plane color coded from faded light green to red. As distinct from the plot, most of the data-points are accumulated at the middle of azimuth and temperature plot. Lowest power is near high azimuth and low temperature, as well as low azimuth and low temperature. The maximum power values seem to be when both parameters show high readings.

Table 4. Symbol Nomenclature.

|   | Parameter | Description |
|---|-----------|-------------|
| 1 | $\alpha$ | Elevation angle |
| 2 | $\beta$ | Tilt angle |
| 3 | $\delta$ | Declination angle |
| 4 | $\phi$ | Latitude of the SPV site |
| 5 | $S_i$ | Solar radiation incident perpendicular to the sun (DR_Avg) |
| 6 | $S_m$ | Solar radiation incident on an inclined surface |
| 7 | $S_h$ | Solar radiation on a horizontal surface |
| 8 | $P_c$ | Calculated power of the module |
| 9 | $w_i$ | Incident wind speed on the panel (WS_Avg) |
| 10 | $T_i$ | Temperature of the PV module changing due to wind speed |
| 11 | $k1 - k6$ | Constants for the PV module |
| 12 | $P_r$ | Packing ratio |

Table 5. Power Equation Parameter values.

| Parameter | Value |
|-----------|-------|
| Area | $21.77\ m^2$ |
| Efficiency | 0.19 |
| $P_r$ = Efficiency*Area | 12.36 |
| $\phi$ | $26.2°$ |
| $\beta$ | $26°$ |
| $\alpha = 90 - \phi + \beta$ | $89.8°$ |
| k1 | -0.06689 |
| k2 | -0.012844 |
| k3 | -0.002262 |
| k4 | 0.0002276 |
| k5 | 0.000159 |
| k6 | -0.000006 |

Table 6. Actual vs Calculated power.

|   | Timestamp | Power Actual | Power Calculated |
|---|-----------|-------------|------------------|
| 1 | 2019-01-05 09:17:00 | 3355.04 | 2808.03 |
| 2 | 2019-01-05 09:18:00 | 3346.00 | 2766.82 |
| 3 | 2019-01-05 09:19:00 | 3284.97 | 2768.39 |
| 4 | 2019-01-05 09:20:00 | 3426.53 | 2800.90 |
| 5 | 2019-01-05 09:21:00 | 3324.33 | 2835.44 |
| 6 | 2019-01-05 09:22:00 | 3480.50 | 2847.67 |
| 7 | 2019-01-05 09:23:00 | 3432.44 | 2894.21 |
| 8 | 2019-01-05 09:24:00 | 3406.01 | 2847.34 |
| 9 | 2019-01-05 09:25:00 | 3487.33 | 2869.64 |
| 10 | 2019-01-05 09:26:00 | 3554.86 | 2918.95 |
| 11 | 2019-01-05 09:27:00 | 3637.89 | 2975.86 |
| 12 | 2019-01-05 09:28:00 | 3606.71 | 2979.75 |
| 13 | 2019-01-05 09:29:00 | 3602.64 | 3034.08 |
| 14 | 2019-01-05 09:30:00 | 3665.04 | 3038.76 |
| 15 | 2019-01-05 09:31:00 | 3678.98 | 3074.13 |
| 16 | 2019-01-05 09:32:00 | 3868.26 | 3165.09 |
| 17 | 2019-01-05 09:33:00 | 3910.53 | 3216.70 |
| 18 | 2019-01-05 09:34:00 | 3906.38 | 3210.01 |
| 19 | 2019-01-05 09:35:00 | 3870.09 | 3250.37 |
| 20 | 2019-01-05 09:36:00 | 3983.11 | 3261.65 |

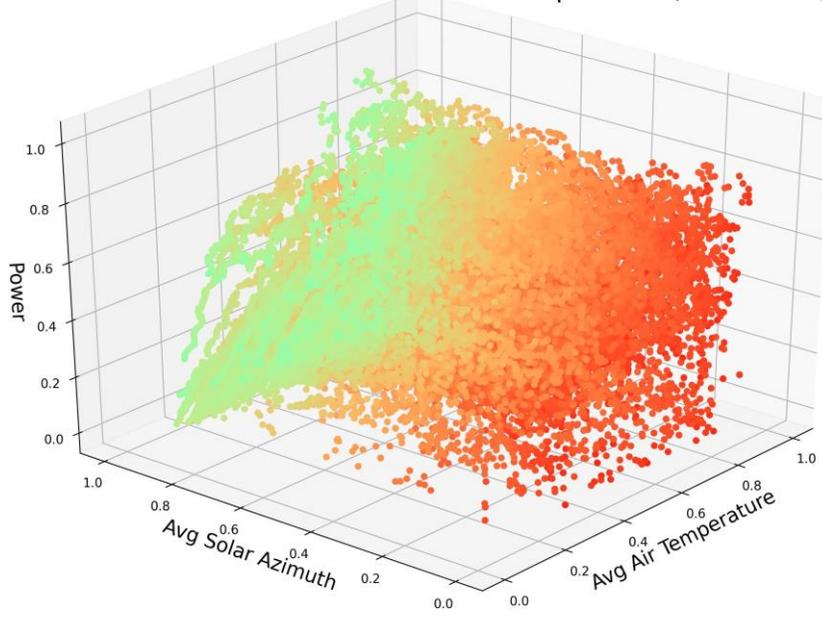

Power Vs Solar Azimuth and Ambient Temperature (normalized).

Fig. 11. Visualization of variation of power measurement with respect to solar azimuth angle and ambient temperature measurements.

## 4 ENSEMBLE LEARNING MODELS EXPLORATION, EXPERIMENTATION AND RESULT ANALYSIS

### 4.1 Model Exploration

As explained by Buhlmann *et al*. [37], in case of estimation of a function expressed as:

$$g : \mathbb{R}^d \leftarrow \mathbb{R} \qquad (5)$$

where d is the dimension of the predictor variable, the ensemble function estimate is given as:

$$\hat{g}_{ens}(\cdot) = \sum_{k=1}^{M} c_k \hat{g}_k (\cdot) \qquad (6)$$

where M is the number of function estimates got by using re-weighted input dataset. In this paper, the following ensemble learning methods have been implemented:-

#### 4.1.1 *Bagging:* For a given number of *base procedures*, bagging based smoothing operation reduces the variance and hence the mean square error. If $X_i \in \mathbb{R}^d$ is the $d$-dimensional predictor variable and $Y_i \in \mathbb{R}$, where $i \in (1, n)$ for n data points, then the function estimator can be given as:

$$\hat{g}(\cdot) = h_n\big((X_1, Y_1), (X_2, Y_2), ...., (X_n, Y_n)\big)(\cdot) : \mathbb{R}^d \leftarrow \mathbb{R} \qquad (7)$$

A bagging estimator can be formulated as follows:

$$\hat{g}_{Bag}(\cdot) = \mathbb{E}^*\big[\hat{g} * (\cdot)\big] \qquad (8)$$

**Random Forest Regression:** Random Forest introduced by Breiman *et al*. [38], is a tree based ensemble where each tree structure depends on a collection of random variables. These trees represented as $h_j(X, \theta_j)$ are called *base learners*. For a certain set of data $D = (x_1, y_1), (x_N, y_N)$ and a particular realization $\theta_j$ of $\Theta_j$, the fitted tree is represented as $\hat{h}_j(x, \theta_j, D)$. Considering a joint distribution $\mathrm{P}_{XY}(X, Y)$, where $X$ and $Y$ are vectors containing predictor and response variables respectively, then the prediction function $f(X)$ is determined by a loss function $L(Y, f(X))$ tries to minimize the expected value of the loss $\mathbb{E}_{XY}(L(Y, f(X)))$ and hence the regression function can be coined as $f(X) = \mathbb{E}(Y|X = x)$. In regression task of random forest, the *base learners* $h_1(x), ...h_j(x)$ constructing the ensemble f are averaged as

$$f(x) = 1/J \sum_{j=1}^{J} h_j(x)$$

**Extreme Randomized Trees Regression:** Geurts *et al.* [39] proposed an algorithm that builds an ensemble of un-pruned decision trees in a top-down manner. As a major distinguishing feature, it splits the nodes of the trees by randomly choosing cut-points fully and that instead of a bootstrap replica it uses the whole learning sample to grow the trees. Extra-Trees splitting algorithm is used multiple times with the learning sample (the observations used to build a model) to finally generate an ensemble model. There are a number of parameters that determine the performance of extra-trees, for example, K (number of attributes randomly selected at each node) gives an estimate on the strength of the sample selection process, $n_{min}$ (minimum sample si for splitting a node) provides an idea on the strength of averaging the output noise, and $M$ the strength of the variance reduction of the ensemble model aggregation

*4.1.2* *Boosting:* In boosting method of ensemble the *base procedures* operate sequentially. Boosting is non-parametric optimization algorithm often considered better performing than bagging algorithm for regression, classification, etc. Based on the data represented as $(X_i, Y_i) \forall i \in (1, n)$, where $n$ is the number of data-points, boosting algorithms estimate the function $g: R_d \rightarrow R$ and minimize the expected loss $E[l(Y, g(X)], l(\cdot, \cdot): R \times R \rightarrow R^+$. Here, $Y$ can be both continuous and discrete for regression and classification tasks respectively.

**AdaBoost:** Freund *et al.* [40, 41] introduced the Ada (adaptive) Boost algorithm that attempts to reduce error of ensemble by consistently generating classifiers. The algorithm aims to reach a hypothesis while minimizing the error of a distribution $D$ over training samples $(x, y)$. The algorithm trains learners iteratively by normalizing a set of weights at $t_{th}$ iteration such that the distribution $P_t$ calculated from this are fed into the weak *base learners*. These learners finally generate a hypothesis $h_f$ with minimal error in the distribution. The hypothesis can be formulated as:

$$h_f(x) = \begin{cases} 1, & if \sum_{t=1}^{T} (log \; 1/B_t) h_t(x) \geq 1/2 \sum_{t=1}^{T} log \; 1/B_t \\ 0, & otherwise \end{cases}$$

where the $B$ is a parameter to update the weight vector and $h_f$ is the combined hypothesis of the outputs of the $T$ weak hypotheses based on a weighted majority vote.

**Gradient Boosting Machine (GBM):** Friedman *et al.* [42] proposed the stochastic gradient boosting algorithm that aims to find a function F_*(x) mapping $x$ with $y$ (input variable and output variable respectively) so that the expected loss over the joint distribution of $(x, y)$, given by $\Psi(y, F(x))$ is minimized. The minimization function can be written as:

$$F_*(x) = \underset{F(x)}{\arg \min} \; E_{(y,x)} \Psi(y, F(x)) \tag{9}$$

The major concept adopted in such algorithm is to build new weak base learners with maximum correlation with the negative gradient of the loss function mentioned above. These new models or base learners are sequentially stacked on top to generate the final ensemble. The error or loss is calculated on the ensemble to train each learner. Gradient boosting machines are known for their flexibility of configuration, since out of a variety of loss functions one can be chosen that can work on training the ensemble. Hence, the algorithm can be easily customized to any data driven task [43].

**XGBoost:** XGBoost by Chen *et al.* [44], is a scalable tree boosting sparsity aware algorithm. In this boosting algorithm the predictive model is trained in additive manner. At $t_{th}$ iteration and for prediction of $i_{th}$ data instance, $f_t$ is added to minimize the objective function: $L^{(t)} = \sum_{i=1}^{n} l(y_i, y_i^{(t-1)} + f_t(x_i)) + \Omega(f_t)$, where $\Omega$ penalizes complexity of the model in XGBoost, $l$ is a differentiable convex loss function that measures the difference between the prediction $\hat{y}_i$ and the target $y_i$, and $x_i$ is the input.

**LightGBM:** LightGBM [45] is a gradient boosting framework based on gradient boosting decision trees (GBDT) with gradient based one side sampling (GOSS) and exclusive feature bundling (EFB), implemented to increase the efficiency of the algorithm while reducing memory usage and training time. GBDT through decision trees learns a function mapping the input space $X_s$ to gradient space $G_s$. For each feature i, GBDT determines $d^*_j = \arg\max_d V_j(d)$ while calculating the largest information gain $V_j(d^*_j)$, where $V$ is the variance gain of splitting the feature j at data-point d for a fixed node of the decision tree. However, in GOSS, an estimated $V_j(d)$ over a smaller data instance subset, thus reducing the computational cost immensely.

**HGBoost:** Histogram based algorithm maps unsorted original data values to discrete values to index the histogram, thus transforming floating-point data-type to character as binned data. For XGBoost, original data values are pre-sorted thus both data and their indices need to be save. Histogram-based boosting algorithm [46] is such that the ensemble's memory cost is 8 times lower than XGBoost for the same dataset, as well as the computational cost while training is lower. HGBoost regression estimator has native support for missing values in dataset. This is modeled from the LightGBM, which originally introduced the idea of histogram algorithm.

*4.1.3 Voting-averaged Regressor Ensemble:* An et al. [47] introduced the voting average scheme for regression ensembles. For a regression function $f = (x_1; y_1), (x_2; y_2), \dots (x_t; y_t)$ there are $n$ base learners $R_i(i = 1, \dots, n)$ with t training samples. A simple weighted average for such ensemble is $\hat{y}_j = \sum_{i=1}^n w_i x_{ij} = \sum_{i=1}^n w_i R_i(x_j), \forall j \in (1, t)$, where $wi$ is the weight of the ith regressor Ri, $\hat{y}$ is the ensemble output, and $z_{ij}$ is the i-th regressor output corresponding to the $j$-th training sample $x_i$, $z_{ij} = R_i(x_j)$. On top of this, the voting average ensemble does classification of the regressor output into $m$ classes $C_{jk} \forall k (1, m)$, a majority voting where the the $k*$-th class is the winner class such as $C_{jk*} = \text{Voting}(C_{j1}, \dots C_{jm})$, an averaging of the $k*$-th class corresponding to the $j$-th sample given as $\hat{y}_j = Averaging(C_{jk*})$.

*4.1.4 Stacked Regression Ensemble:* Breinman *et al.* [48] introduced the notion of linearly combining predictors to use cross-validation data and least squares under non-negative constraint. For $K$ regressors in the ensemble, the learning set can be represented as $\mathbb{L} = (y_n, x_n), n = 1, 2, \dots N$ where $x$ and $y$ are input and predicted vectors respectively. The problem for calculating the predicted vectors is simplified by restriction attention to combinations given as $v(x) = \sum_k a_k v_k(x)$ such that for $\mathbb{L}$ one possibility is to take the $\alpha_k$ to minimize the expression $\sum_n (y_n - \sum_k a_k v_k(x_n))^2$.

It is experimentally proved in earlier works that stacking a number of regressions can produce decreased error rates.

## 4.2 Experimentation and Evaluation

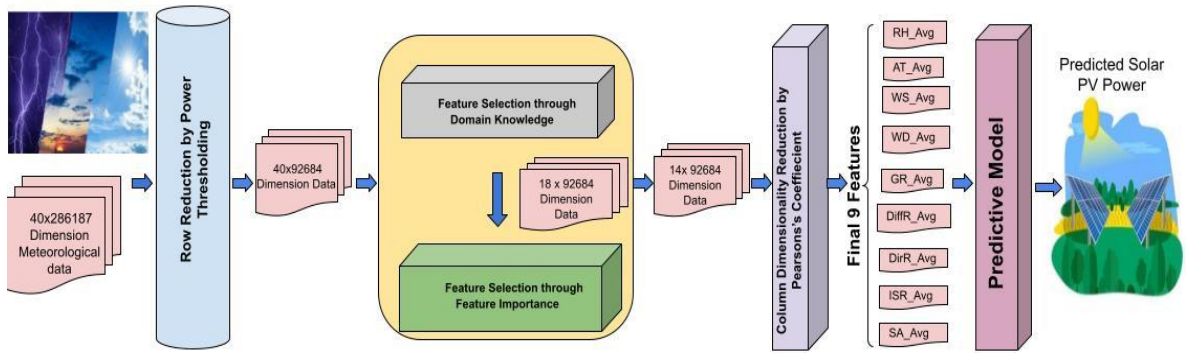

Fig. 12. Generic prediction framework used for all the learning models implemented in this paper.

In this section of the paper, the various steps involved in training the ML models and evaluation metric used to compare the various ML models have been explained. The prediction framework has been represented in Figure 12.

- **Train-Test Split :**

  The next step after data pre-processing and analysis, is to train different regression models relevant to the dataset. The cleaned and reduced data is split into training set with 80 % of data for training the predictive models. The rest 20 % is used for testing the model.

- **Evaluation Metrics:**

  To evaluate model's performance, some evaluation metrics have been defined which are root mean squared error (RMSE) and R-Square score ($R^2$). The prediction framework has been represented in Figure 12.

$$RMSE = \frac{\sqrt{\sum_{n=1}^{N}(x_i - \hat{x_i})^2}}{N} \tag{10}$$

$$R^2 = \frac{SS_{regression}}{SS_{total}} \tag{11}$$

- **Experimentation:**

In the proposed prediction test-bed SRRA dataset has been cleaned with appropriate approaches and analysis has been done which help in better prediction accuracy for the models. These set of selected features areused for training all the classical ML and EML models. To set a benchmark for evaluation, classical ML models like linear regression, elastic regression, bayesian ridge regression and K-nearest neighbor have been trained. Hyper-parameters has been tuned to get the best performance of the models. The RMSE plot for the same has been given in Figure 13(a). The selected set of classical models were able to predict the power generated and served as a reasonable benchmark for the next set of model training.

In [17], [49] and [18] the authors used ensemble ML technique for short term prediction of solar irradiation for PV generators. In [49] authors have provided a review and categorization on the ensemble methods for wind speed/solar irradiance forecasting. They provided a comparison of 4 ensemble methods for solar irradiance prediction, on subsets of the time-series data from NSRDB datasets in terms of root mean square error and mean absolute scaled error. Although these approaches have proposed use of ensemble mechanisms but they do not involve extensive analysis and processing on meteorological data. To validate the effect of processed region specific test dataset, training and testing on few ensemble techniques like Histogram Gradient Boosting, XGBoost, Gradient Boosting, Light Gradient Boosted Machine, Extra-Trees, Random Forest, AdaBoost has been performed. The hyper-parameters have been tuned to give best results and have been listed in Table 7. The RMSE of the ensemble models trained have been noted in Table 8. Further, top 4 scoring models (histogram gradient boosting, light gradient boosted machine, extra-trees, and random forest), pipeline for voting and stacking regressor are developed. In a similar manner stacking and voting ensemble algorithms have also been trained and tested. The time taken for training and testing of the models have been listed in Table 9.

The proposed test bed is developed in python language on a standard laptop computer with the specifications asstated in Table 10. **Sklearn** machine learning libraries[5] are used for our experimentations.

Table 7. Hyperparameters tuning.

| Model | Hyperparameters for ML models |
|---|---|
| Histogram Gradient Boosting Regression | learning_rate=0.02, loss = 'least_absolute_deviation', max_depth = 40, max_iter = 750, min_samples_leaf = 2, random_state = 73 |
| Light Gradient Boosted Machine | objective = 'regression', num_leaves = 5, learning_rate = 0.05, n_estimators = 720, max_bin = 55, bagging_fraction = 0.8, bagging_freq = 5, feature_fraction = 0.2319, feature_fraction_seed = 9, bagging_seed = 9, min_data_in_leaf = 6, min_sum_hessian_in_leaf = 11 |
| Extra-Trees Regressor | n_estimators = 1000, random_state = 73 |
| Random Forest Regressor | bootstrap = False, max_depth = 60, max_features = 'sqrt', min_samples_split = 4, n_estimators = 1700 |
| XGBoost Regression | n_estimators = 1200, learning_rate = 0.05, random_state = 73 |
| Gradient Boosting Regressor | max_depth = 10, max_features = 'sqrt', min_samples_split = 12, n_estimators = 1500 |
| AdaBoost Regressor | learning_rate = 0.03, loss = 'exponential', n_estimators = 2300, random_state = 73 |

Table 8. RMSE and R2 score comparison table.

| Sr.No. | Ensemble Technique | RMSE ↓ | R2 Score ↑ |
|---|---|---|---|
| 1 | Voting Regressor | **313.07** | **0.96** |
| 2 | Stacking Regressor | **314.90** | **0.96** |
| 3 | Histogram GradientBoosting Regressor | 345.64 | 0.94 |
| 4 | LightGradient Boosting Machine | 363.64 | 0.94 |
| 5 | Extra-Trees Regressor | 389.44 | 0.93 |
| 6 | Random Forest Regressor | 398.62 | 0.93 |
| 7 | XGBoost Regressor | 433.42 | 0.93 |
| 8 | GradientBoostingRegressor | 540.34 | 0.91 |
| 9 | Ada-Boost Regressor | 1139.38 | 0.526 |

Table 9. Train-test time table.

| Ensemble Technique | Training time taken (seconds) | Avg. time required per prediction (micro-seconds) |
|---|---|---|
| Voting Regressor | 192.66 | 115.11 |
| Stacking Regressor | 981.60 | 114.5 |
| Random Forest Regressor | 517.63 | 567.50 |
| Extra-Trees Regressor | 157.91 | 565.30 |
| Ada-Boost Regressor | 229.97 | 226.00 |
| Gradient Boosting Regressor | 310.02 | 2.98 |
| Histogram Gradient Boosting Regressor | 10.17 | 5.12 |
| XGBoost Regressor | 48.16 | 3.33 |
| Light Gradient Boosting Machine | 1.26 | 2.15 |

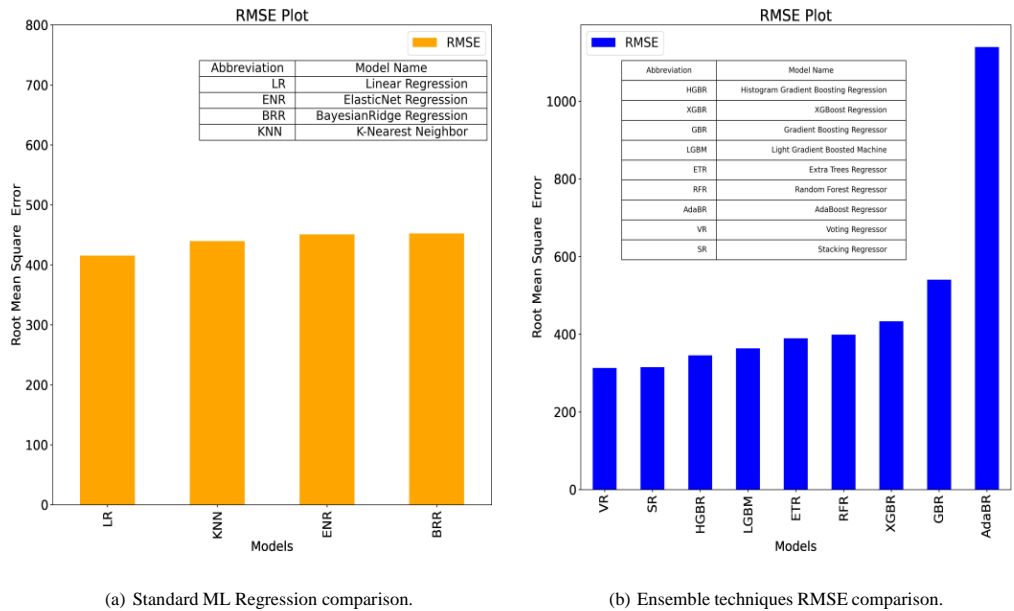

(a) Standard ML Regression comparison.

(b) Ensemble techniques RMSE comparison.

Fig. 13. Models vs. RMSE plots.

Table 10. System configurations used.

| Processor | Speed | Memory | OS |
|---|---|---|---|
| Ryzen 7 4800 CPU | 2.90 GHz | 16 GB | Windows 10 64 bit |

### 4.3 Result Analysis and Discussions for Future

The various aspects of the proposed work, its results, and outcome have been discussed and summarized in this section.

*4.3.1 Analysis of Test Bed and Ensemble Techniques.* Classical ML models like linear regression, elastic net regression, Bayesian ridge regression and k-nearest neighbor set a good benchmark of prediction accuracy, given in Figure 13(a). It was observed from the results that linear regression gave the best result with RMSE of around 415. In this work EML algorithms that are simple to implement have been reviewed with acceptable range of error. From Figure 13(b), it is found that most EML algorithms were better than the benchmark set by linear regression except for gradient boosting, xgboost and adaboost regressors. It was also analyzed that the EML algorithms mostly underestimated the power output and hence helps to get the minimum expected power generated from the point of view of power plant establishment. With reference to Figure 13(b) it is observed that voting and stacking regressors has the best prediction accuracy with R2 score of 0.96 and RMSE scores of 313.07 and 314.90 respectively. According to all the error criteria used, voting and stacking regressors are found to be more suited for Eastern India region. Average of 20 set of accuracy observations have been considered for model comparison.

To analyze the prediction plots, an interval of dataset has been randomly sampled and represented it scatter plots of actual, predicted power vs observation count (Figures 14 and 15 ). The same interval is used as an example to compare all the classical ML algorithms and EML algorithms. This helps us to visualize the nature of deviations of predicted power from the actual power. Referring to Figures 14(a), 14(b) and 14(c) it is observed that linear regression, Bayesian ridge regression and elasticnet regression does a good job in terms of predicting the actual power but KNN regression 14(d) has lots of difference in actual vs predicted observations. It is clearly evident from Figure 15(a),Figure 15(b) and the accuracy in Figure 13(b) that voting and stacking algorithms perform much better than the classical ML algorithms and adaboost regressor in Figure 15(h) performs the worst among all other algorithms. Figures 16 and 17 demonstrates the results for actual and predicted power vs number of observations over several days.

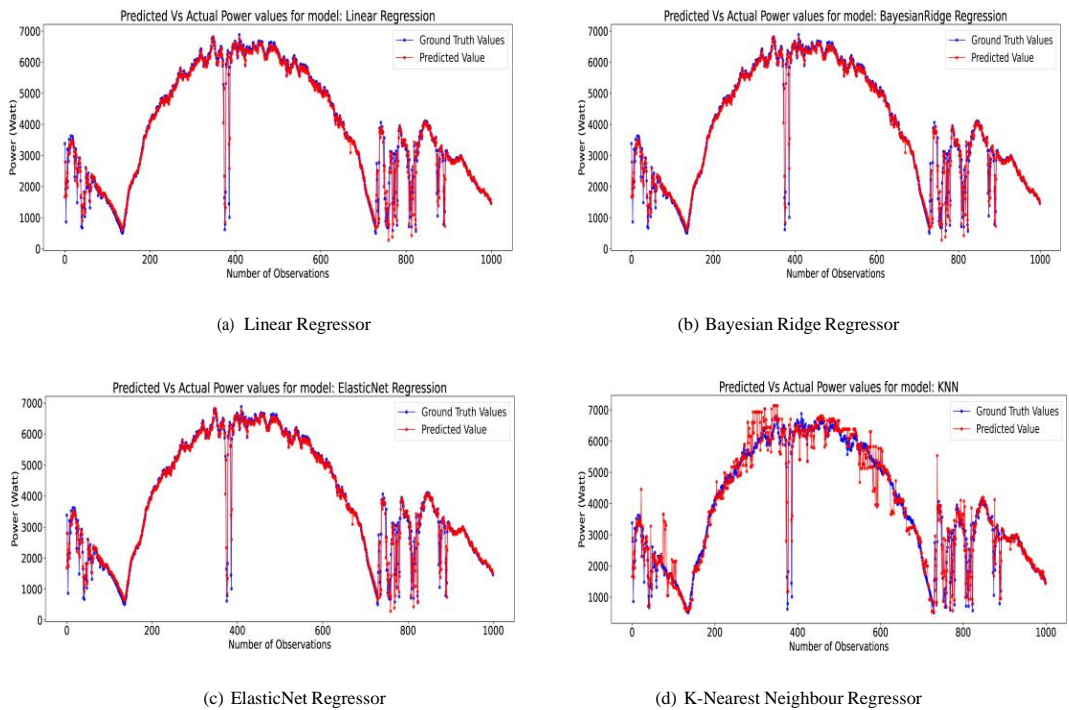

(a)  Linear Regressor

(b)  Bayesian Ridge Regressor

(c)  ElasticNet Regressor

(d)  K-Nearest Neighbour Regressor

Fig. 14. Random Sample plot of Actual and Predicted Energy over number of observations for ML models.

*4.3.2    Future scope and research direction.* In this work the authors have performed extensive survey and validation on wide collection of ensemble learning methods which claims to be very useful for optimization of solar power plant under dynamic weather conditions. The current research team is subsequently focusing on other ML algorithms (transfer learning, meta learning, domain adaptation, reinforcement learning) and development of new algorithms and loss functions for optimizing algorithm performance. The current explored work can also be implemented for estimation of Return on Investment (ROI) for region based large scale solar PV power plants.

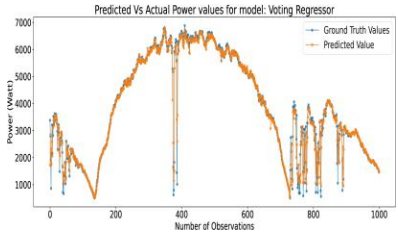
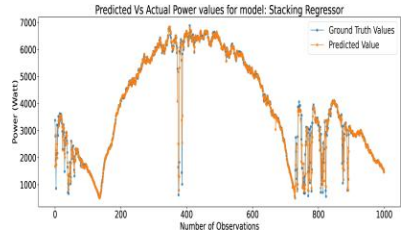

(a) Voting Regressor          (b) Stacking Regressor

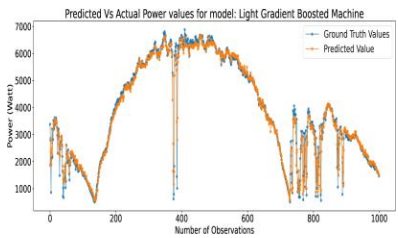
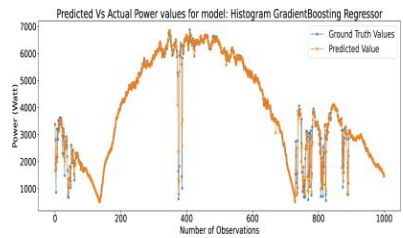

(c) Light Gradient Boosted Machine     (d) Histogram Gradient Boosting Regressor

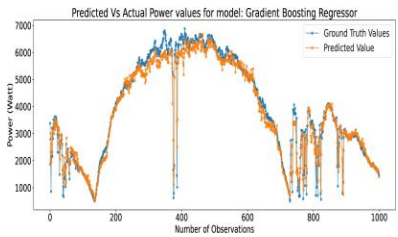
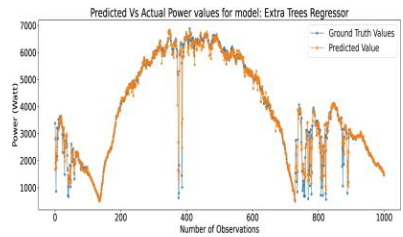

(e) Gradient Boosting Regressor (f) Extra-Trees Regressor

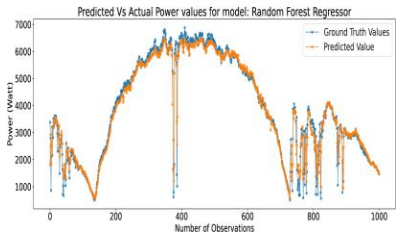
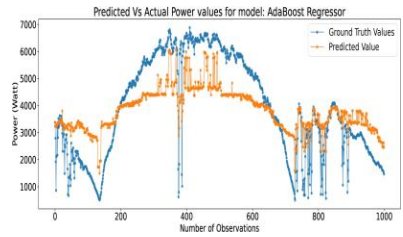

(g) Random Forest Regressor     (h) AdaBoost Regressor

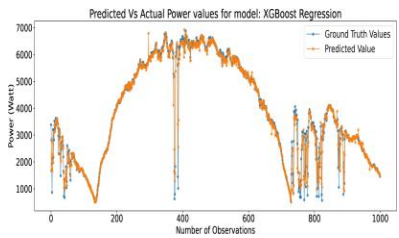

(i) XGBoost Regressor

Fig. 15. Random sample plot of Actual and Predicted Energy over number of observations.

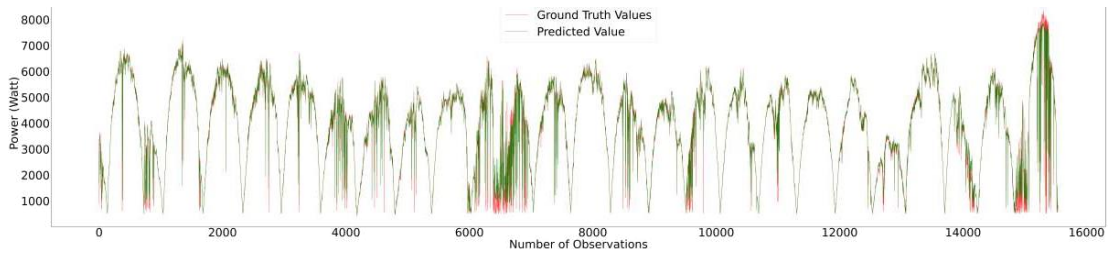

Fig. 16. Actual and predicted power vs number of observations over several days for Voting Regressor.

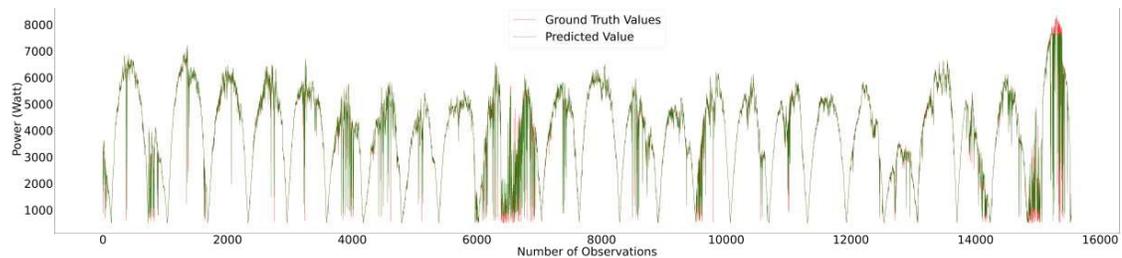

Fig. 17. Actual and predicted power vs number of observations over several days for Stacking Regressor.

## 5 CONCLUSION

In this paper, various ensemble machine learning algorithms have been compared for prediction of solar power generation under the impact of meteorological data. Although there are a good number of existing reports on solar power prediction using traditional deduction methods, machine learning approaches or deep learning based frameworks,but there does not exist a comprehensive case study on regional solar power generation data proposing end-to-end solution from data preparation to machine learning models performance comparison. This is very critical to ensure thatseveral geographical and topological factors that can influence the weather parameters directly and thereby the solar power generation in the region of interest are considered while training the prediction models. With the case study report shown in this paper, the greenfield solar projects in the eastern region of India can refer the comparison analysisof various EML models to choose the most convenient prediction model. A test-bed for learning models has been designed, to collect, curate, and analyze dataset as well as select appropriate learning models based on meteorological parameters. The framework also supports feature selection and dimensionality reduction for dataset to reduce space andtime complexity of the learning models. The prediction results obtained in this work demonstrates that the performanceaccuracy of voting and stacking algorithm gives the best results (313.07 RMSE, 0.96 R2-score and 314.9 RMSE, 0.96 R2-score respectively). The proposed EML based prediction work is a generalized one thus can be very useful for predicting the performance of even large-scale solar PV power plants considering similar meteorological conditions. The presented work can be subsequently extended to techno-financial analysis (such as plant efficiency, ROI) of already running and newly installed solar power plants considering various efficient deep learning algorithms. Further study can be done for optimization of ML algorithm performance and for designing improved prediction algorithms.

## 6 ACKNOWLEDGEMENT

The authors are thankful to Prof. Hiranmay Saha, for providing with the facility of SRRA Centre for obtaining the real weather data and the solar PV power plant, installed at the roof top of the CEGESS, IIEST, Shibpur, funded by the Ministry of Renewable Energy (MNRE), Govt. of India. The research work has been done at BITS, Pilani, Hyderabad Campus.

## REFERENCES

[1] U.S. Energy Information Administration. https://www.eia.gov/outlooks/steo/, 2022.

[2] India Brand Equity Foundation. Renewable energy. https://www.ibef.org/industry/renewable-energy.aspx, 2021.

[3] Jan Kleissl. *Solar energy forecasting and resource assessment*. Academic Press, 2013.

[4] Hassan Ismail Fawaz, Germain Forestier, Jonathan Weber, Lhassane Idoumghar, and Pierre-Alain Muller. Deep learning for time series classification: a review. *Data mining and knowledge discovery*, 33(4):917–963, 2019, doi:https://doi.org/10.1007/s10618-019-00619-1.

[5] Emilio Pérez, Javier Pérez, Jorge Segarra-Tamarit, and Hector Beltran. A deep learning model for intra-day forecasting of solar irradiance using satellite-based estimations in the vicinity of a pv power plant. *Solar Energy*, 218:652–660, 2021, doi:10.1016/j.solener.2021.02.033.

[6] Jae-Ju Song, Yoon-Su Jeong, and Sang-Ho Lee. Analysis of prediction model for solar power generation. *Journal of digital convergence*, 12(3):243–248, 2014, doi:10.14400/JDC.2014.12.3.243.

[7] Francisco J Rodríguez-Benítez, Clara Arbizu-Barrena, Javier Huertas-Tato, Ricardo Aler-Mur, Inés Galván-León, and David Pozo-Vázquez. A short-term solar radiation forecasting system for the iberian peninsula. part 1: Models description and performance assessment. *Solar Energy*, 195:396–412, 2020, doi:https://doi.org/10.1016/j.solener.2019.11.028.

[8] Javier Huertas-Tato, Ricardo Aler, Inés M Galván, Francisco J Rodríguez-Benítez, Clara Arbizu-Barrena, and David Pozo-Vázquez. A short-term solar radiation forecasting system for the iberian peninsula. part 2: Model blending approaches based on machine learning. *Solar Energy*, 195:685–696, 2020, doi:https://doi.org/10.1016/j.solener.2019.11.091.

[9] Fermín Rodríguez, Ainhoa Galarza, Juan C Vasquez, and Josep M Guerrero. Using deep learning and meteorological parametersto forecast the photovoltaic generators intra-hour output power interval for smart grid control. *Energy*, 239:122116, 2022, doi:https://doi.org/10.1016/j.energy.2021.122116.

[10] Takeshi Watanabe, Hideaki Takenaka, and Daisuke Nohara. Post-processing correction method for surface solar irradiance forecast data from the numerical weather model using geostationary satellite observation data. *Solar Energy*, 223:202–216, 2021, doi:https://doi.org/10.1016/j.solener.2021.05.055.

[11] Hadrien Verbois, Robert Huva, Andrivo Rusydi, and Wilfred Walsh. Solar irradiance forecasting in the tropics using numerical weather prediction and statistical learning. *Solar Energy*, 162:265–277, 2018, doi:https://doi.org/10.1016/j.solener.2018.01.007.

[12] Da Liu and Kun Sun. Random forest solar power forecast based on classification optimization. *Energy*, 187:115940, 2019, doi:https://doi.org/10.1016/j.energy.2019.115940.

[13] Xing Luo, Dongxiao Zhang, and Xu Zhu. Deep learning based forecasting of photovoltaic power generation by incorporating domain knowledge. *Energy*, 225:120240, 2021, doi:https://doi.org/10.1016/j.energy.2021.120240.

[14] Amir Rafati, Mahmood Joorabian, Elaheh Mashhour, and Hamid Reza Shaker. High dimensional very short-term solar power forecasting based on a data-driven heuristic method. *Energy*, 219:119647, 2021, doi:https://doi.org/10.1016/j.energy.2020.119647.

[15] Imane Jebli, Fatima-Zahra Belouadha, Mohammed Issam Kabbaj, and Amine Tilioua. Prediction of solar energy guided by pearson correlation using machine learning. *Energy*, 224:120109, 2021, doi:https://doi.org/10.1016/j.energy.2021.120109.

[16] Herman Böök and Anders V Lindfors. Site-specific adjustment of a nwp-based photovoltaic production forecast. *Solar Energy*, 211:779–788, 2020, doi:https://doi.org/10.1016/j.solener.2020.10.024.

[17] Waqas Khan, Shalika Walker, and Wim Zeiler. Improved solar photovoltaic energy generation forecast using deep learning-based ensemble stacking approach. *Energy*, 240:122812, 2022, doi:https://doi.org/10.1016/j.energy.2021.122812.

[18] Fermín Rodríguez, Fernando Martín, Luis Fontán, and Ainhoa Galarza. Ensemble of machine learning and spatiotemporal parameters to forecast very short-term solar irradiation to compute photovoltaic generators' output power. *Energy*, 229:120647, 2021, doi:https://doi.org/10.1016/j.energy.2021.120647.

[19] Alexis Fouilloy, Cyril Voyant, Gilles Notton, Fabrice Motte, Christophe Paoli, Marie-Laure Nivet, Emmanuel Guillot, and Jean-Laurent Duchaud. Solar irradiation prediction with machine learning: Forecasting models selection method depending on weather variability. *Energy*, 165:620–629, 2018, doi:https://doi.org/10.1016/j.energy.2018.09.116.

[20] Peder Bacher, Henrik Madsen, and Henrik Aalborg Nielsen. Online short-term solar power forecasting. *Solar Energy*, 83(10):1772–1783, 2009, doi:10.1016/j.solener.2009.05.016.

[21] Federica Davò, Stefano Alessandrini, Simone Sperati, Luca Delle Monache, Davide Airoldi, and Maria T Vespucci. Post-processing techniques and principal component analysis for regional wind power and solar irradiance forecasting. *Solar Energy*, 134:327–338, 2016, doi:https://doi.org/10.1016/j.solener.2016.04.049.

[22] Jie Shi, Wei-Jen Lee, Yongqian Liu, Yongping Yang, and Peng Wang. Forecasting power output of photovoltaic systems based on weather classification and support vector machines. *IEEE Transactions on Industry Applications*, 48(3):1064–1069, 2012, doi:10.1109/IAS.2011.6074294.

[23] Ricardo Aler, Ricardo Martín, José M. Valls, and Inés M. Galván. A study of machine learning techniques for daily solar energy forecasting using numerical weather models. In David Camacho, Lars Braubach, Salvatore Venticinque, and Costin Badica, editors, *Intelligent Distributed Computing VIII*, pages 269–278, Cham, 2015. Springer International Publishing.

[24] Junliang Fan, Xiukang Wang, Lifeng Wu, Hanmi Zhou, Fucang Zhang, Xiang Yu, Xianghui Lu, and Youzhen Xiang. Comparison of support vector machine and extreme gradient boosting for predicting daily global solar radiation using temperature and precipitation in humid subtropical climates: A case study in china. *Energy conversion and management*, 164:102–111, 2018, doi:https://doi.org/10.1016/j.enconman.2018.02.087.

[25] Rachit Srivastava, AN Tiwari, and VK Giri. Forecasting of solar radiation in india using various ann models. In *2018 5th IEEE Uttar Pradesh section international conference on electrical, electronics and computer engineering (UPCON)*, pages 1–6. IEEE, 2018.

[26] Phathutshedzo Mpfumali, Caston Sigauke, Alphonce Bere, and Sophie Mulaudzi. Day ahead hourly global horizontal irradiance forecasting—application to south african data. *Energies*, 12(18):3569, 2019, doi:https://doi.org/10.3390/en12183569.

[27] Veysel Çoban and Sezi Çevik Onar. Solar radiation prediction based on machine learning for istanbul in turkey. In *International Conference on Intelligent and Fuzzy Systems*, pages 197–204. Springer, 2019.

[28] Zezhou Chen and Irena Koprinska. Ensemble methods for solar power forecasting. In *2020 International Joint Conference on Neural Networks (IJCNN)*, pages 1–8. IEEE, 2020.

[29] PAGM Amarasinghe, NS Abeygunawardana, TN Jayasekara, EAJP Edirisinghe, and SK Abeygunawardane. Ensemble models for solar power forecasting—a weather classification approach. *AIMS Energy*, 8(2):252–271, 2020, doi:10.3934/energy.2020.2.252.

[30] Tendani Mutavhatsindi, Caston Sigauke, and Rendani Mbuvha. Forecasting hourly global horizontal solar irradiance in south africa using machine learning models. *IEEE Access*, 8:198872–198885, 2020, doi:10.1109/ACCESS.2020.3034690.

[31] Mawloud Guermoui, Farid Melgani, Kacem Gairaa, and Mohamed Lamine Mekhalfi. A comprehensive review of hybrid models for solar radiation forecasting. *Journal of Cleaner Production*, 258:120357, 2020, doi:https://doi.org/10.1016/j.jclepro.2020.120357.

[32] Yong Zhou, Yanfeng Liu, Dengjia Wang, Xiaolun Liu, and Yingying Wang. A review on global solar radiation prediction with machine learning models in a comprehensive perspective. *Energy Conversion and Management*, 235:113960, 2021, doi:https://doi.org/10.1016/j.enconman.2021.113960.

[33] Imane Jebli, Fatima-Zahra Belouadha, Mohammed Issam Kabbaj, and Amine Tilioua. Deep learning based models for solar energy prediction. *Advances Sci*, 6:349–355, 2021, doi:10.25046/aj060140.

[34] Shichao Zhang, Jilian Zhang, Xiaofeng Zhu, Yongsong Qin, and Chengqi Zhang. Missing value imputation based on data clustering. In *Transactions on computational science I*, pages 128–138. Springer, 2008.

[35] Philip Sedgwick. Pearson's correlation coefficient. *Bmj*, 345, 2012, doi:10.1136/bmj.e4483.

[36] Muhannad Alaraj, Astitva Kumar, Ibrahim Alsaidan, Mohammad Rizwan, and Majid Jamil. Energy production forecasting from solar photovoltaic plants based on meteorological parameters for qassim region, saudi arabia. *IEEE Access*, 9:83241–83251, 2021, doi:10.1109/ACCESS.2021.3087345.

[37] Peter Bühlmann. Bagging, boosting and ensemble methods. In *Handbook of computational statistics*, pages 985–1022. Springer, 2012.

[38] Leo Breiman. Random forests. *Machine learning*, 45(1):5–32, 2001, doi:https://doi.org/10.1023/A:1010933404324.

[39] Pierre Geurts, Damien Ernst, and Louis Wehenkel. Extremely randomized trees. *Machine learning*, 63(1):3–42, 2006, doi:https://doi.org/10.1007/s10994-006-6226-1.

[40] Yoav Freund and Robert E Schapire. A decision-theoretic generalization of on-line learning and an application to boosting. *Journal of computer and system sciences*, 55(1):119–139, 1997, doi:https://doi.org/10.1006/jcss.1997.1504.

[41] Yoav Freund and Robert E. Schapire. A short introduction to boosting. In *In Proceedings of the Sixteenth International Joint Conference on Artificial Intelligence*, pages 1401–1406. Morgan Kaufmann, 1999.

[42] Jerome H Friedman. Stochastic gradient boosting. *Computational statistics & data analysis*, 38(4):367–378, 2002.

[43] Alexey Natekin and Alois Knoll. Gradient boosting machines, a tutorial. *Frontiers in neurorobotics*, 7:21, 2013, doi:10.3389/fnbot.2013.00021.

[44] Tianqi Chen and Carlos Guestrin. Xgboost: A scalable tree boosting system. In *Proceedings of the 22nd acm sigkdd international conference on knowledge discovery and data mining*, pages 785–794, 2016.

[45] Guolin Ke, Qi Meng, Thomas Finley, Taifeng Wang, Wei Chen, Weidong Ma, Qiwei Ye, and Tie-Yan Liu. Lightgbm: A highly efficient gradient boosting decision tree. *Advances in neural information processing systems*, 30, 2017.

[46] Wentao Cai, Ruihua Wei, Lihong Xu, and Xiaotao Ding. A method for modelling greenhouse temperature using gradient boost decision tree. *Information Processing In Agriculture*, 9(3):343–354, 2022, doi:10.1016/j.inpa.2021.08.004.

[47] Kun An and Jiang Meng. Voting-averaged combination method for regressor ensemble. In *International Conference on Intelligent Computing*, pages 540–546. Springer, 2010.

[48] L. Breiman. Stacked regressions. *Machine Learning*, 24:49–64, 2004.

[49] Ye Ren, PN Suganthan, and N Srikanth. Ensemble methods for wind and solar power forecasting—a state-of-the-art review. *Renewable and Sustainable Energy Reviews*, 50:82–91, 2015, doi:https://doi.org/10.1016/j.rser.2015.04.081.

[50] Voyant, C., Notton, G., Duchaud, J.-L., Almorox, J., & Yaseen, Z. M. (2020). Solar irradiation prediction intervals based on Box–Cox transformation and univariate representation of periodic autoregressive model. *Renewable Energy Focus*, *33*, 43–53. doi:10.1016/j.ref.2020.04.001.

[51] Ayodele, T. R., Ogunjuyigbe, A. S. O., Amedu, A., & Munda, J. L. (2019). Prediction of global solar irradiation using hybridized k-means and support vector regression algorithms. *Renewable Energy Focus*, *29*, 78–93. doi:10.1016/j.ref.2019.03.003.